\documentclass[letterpaper,10pt,conference]{ieeeconf}

\IEEEoverridecommandlockouts
\usepackage{cite}
\usepackage{amsmath,amssymb,amsfonts}
\usepackage{graphicx}
\usepackage{textcomp}
\usepackage{xcolor}
\usepackage{colortbl}

\usepackage{booktabs}
\usepackage{wrapfig}
\usepackage[above,below]{placeins}
\usepackage{url}

\usepackage{float}
\usepackage{tabularx}
\usepackage{tikz}
\usetikzlibrary{arrows.meta,calc}
\usepackage{arydshln}
\ADLinactivate % Enable dotted rules only in the real-robot results table.

\newenvironment{myitem}{\begin{itemize}}{\end{itemize}}

\newcommand{\namo}{{\ttfamily NAMO}}

\newcommand{\name}{{\ttfamily SPAR}}

\title{\huge
SPAR: Search-trained Push Action Ranker for Efficient Navigation Among Movable Obstacles}

\author{Dhruv Metha Ramesh, Tri-An Nguyen, Junshan Huang, Jingjin Yu, Kostas E. Bekris, Abdeslam Boularias}

\begin{document}

\maketitle
\thispagestyle{empty}
\pagestyle{empty}

\begin{abstract}
Navigation Among Movable Obstacles (\namo) requires a
robot to rearrange objects that block access to its goal.
For a pushing robot, progress may depend on a setup push
that enables a later opening without itself clearing a
passage. The number of candidate sequences can grow
exponentially with sequence length, making physics-based
search costly.
We introduce the Search-trained Push Action Ranker
(\name{}), which learns to prioritize push primitives
from simulated search outcomes. Searches over sequences
of up to two pushes per object provide supervision that
distinguishes immediate openings from setup pushes.
A transformer trained with regression and ranking losses
scores primitives using contact tokens conditioned on scene geometry and reachability
regions. The same model guides simulator evaluations
in \name{} Search and selects actions for direct execution
in \name{} Policy. Both modes solve successive keyhole
problems that open access to neighboring free-space
regions within a \namo{} planner.
Across successful runs on $300$ simulated two-push keyhole
problems, \name{} Search reduces median simulator calls
by $7.0\times$, leading to a $3.8\times$ reduction in
planning time relative to Random Search.
Without retraining, the same ranker guides search on
$600$ keyhole problems which have no solution within two pushes. At a depth limit of five pushes, \name{} Search achieves $50.7\%$ success within $100$ simulator calls, compared with $20.2\%$ for Random Search.
These gains carry over to \namo{} planning: on hard
simulated scenes, \name{} Search achieves $47\%$ success
within $100$ simulator calls, compared with $21\%$ for
Random Search. In real-robot experiments, \name{} Policy
succeeds in $37/40$ trials without a physics-based digital
twin, compared with $33/40$ for \name{} Search and $27/40$
for Random Search.
% Project page: \url{https://anonymous2639.github.io}.
\end{abstract}

% The ICRA template has no printed keyword block.
% navigation among movable obstacles, generative planning, diffusion models, learning for planning

\section{Introduction}
\label{sec:intro}

\begin{figure*}[!t]
\centering
\includegraphics[width=\textwidth]{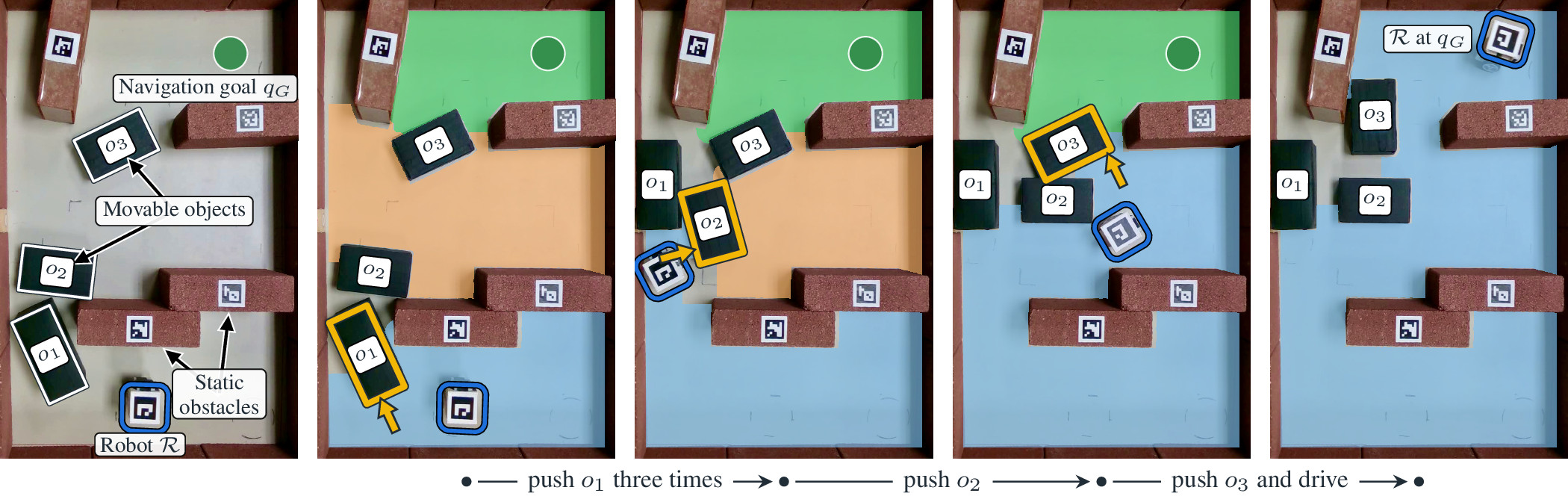}
\caption{Successive keyholes in a hardware \namo\ task. The leftmost
panel labels the robot $\mathcal{R}$, the navigation goal $q_G$, the
movable objects $o_1$, $o_2$, and $o_3$, and static obstacles. In the
remaining panels, from left to right, shading marks the robot's region
(blue), the goal region (green), and a region between them (orange).
Yellow outlines and arrows mark the next pushed object and push
direction, returned by \name{}. Three pushes on $o_1$ expand the robot's region and also
move $o_2$. A push on $o_2$ expands it into the orange region, and a
push on $o_3$ expands it into the goal region, after which
$\mathcal{R}$ drives to $q_G$.}
\label{fig:keyhole_namo}
\end{figure*}

Movable obstacles can easily bring a robot's mission to
a halt. A warehouse inspection robot may need to push boxes aside to continue its route~\cite{ellis2023namo_storagezones}; a robot delivering first aid may need to remove debris to reach a victim~\cite{mueggler2014aerial}. The ability to rearrange such obstacles gives robots a way to create passages through otherwise inaccessible environments.
Navigation Among Movable Obstacles (\namo) captures this
coupling between navigation and obstacle
relocation~\cite{stilman2005namo}. We address \namo{}
for a mobile robot that makes progress toward its goal
through non-prehensile pushing.
Region-based planners organize these rearrangements into
\emph{keyhole problems}, each requiring access to a
neighboring free-space region by moving one or more
blocking objects~\cite{stilman2005namo,he2024interactivefar}.
Each opening extends the robot's access to the workspace
and creates opportunities for further progress. Our focus
is to find useful push sequences for these local problems
with little simulation effort, then compose their solutions
to reach the navigation goal (Fig.~\ref{fig:keyhole_namo}).

% A figure* defined in page-1 text is placed at the top of page 2.

A useful push can matter before it opens a passage.
For example, rotating a box may expose a previously
unreachable face, enabling a second push that clears the
doorway. Finding such \emph{setup pushes} requires accounting for
contact location, push duration, and interactions with
walls or other objects to predict how the object moves
and which contacts the robot can reach
afterward~\cite{stilman2008artificial}.

Physics simulation lets a planner explore how the scene
evolves over successive pushes. However, the number of
candidate sequences can grow exponentially with sequence
length, making even two-push solutions costly to discover.
Prior work learns local pushing
policies~\cite{yao2023local_pushable_rl} or guides sampling
in \namo{} planning~\cite{zhang2025namollm}.
Our Search-trained Push Action Ranker (\name{}) instead
learns from the search itself, turning simulated keyhole
search outcomes into priorities for executable pushes.

Each push primitive specifies a contact point and duration.
We generate training data by searching over sequences
of up to two pushes per object. These searches identify pushes that open a passage immediately and setup
pushes that enable a second push to do so.
Regression and ranking losses train the model to
rank passage-opening pushes above setups, and both above
pushes with no successful continuation.
A transformer uses contact tokens conditioned on object
and obstacle geometry and the source and target free-space
regions to score all primitives for a selected object in
one forward pass, before candidate pushes are simulated.
Because scores depend on the current scene rather than search
depth, the model can be reapplied at each state to guide search
beyond the two-push training horizon.

These priorities support both planning and direct action.
\name{} Search simulates reachable primitives in learned
priority order, retaining alternative sequences when
a branch fails. \name{} Policy instead executes a push
selected from the same scores, then observes and rescores
the changed scene. We evaluate Policy in simulation before
deploying it on a real robot, where it requires no
physics-based digital twin. Both modes use the same
region-graph \namo{} planner, extending the gains on local
keyhole problems to \namo{} without
additional training.

On $300$ simulated two-push keyhole problems, \name{} Search
reduces median simulator calls and planning time over
successful runs by $7.0\times$ and $3.8\times$, respectively,
relative to Random Search. Without retraining, the same ranker guides search on $600$ keyhole problems with no solution within two pushes
under the primitive library. At a depth limit of five
pushes, \name{} Search achieves $50.7\%$ success within
$100$ simulator calls, compared with $20.2\%$ for Random Search.

These gains carry over to \namo{}: on hard simulated
scenes, \name{} Search achieves $47\%$ success within
$100$ simulator calls, compared with $21\%$ for Random Search.
In real-robot experiments, \name{} Policy succeeds in
$37/40$ trials, compared with $33/40$ for \name{} Search
and $27/40$ for Random Search.

Our contributions are:
\begin{myitem}
    \item \name{}, a push ranker learned from simulated
    keyhole searches, combining executable primitives,
    contact-based spatial scoring, and regression and
    ranking supervision that distinguishes immediate
    openings from setup pushes.
    \item Generalization beyond the training horizon:
    priorities learned from searches of up to two pushes
    per object guide search at depths three through five
    on keyhole problems with no solution within two pushes,
    improving success over Random Search at matched
    depths and simulator-call budgets.
    \item Integration of the ranker into search and direct
    execution within a \namo{} planner, with simulation
    and real-robot experiments showing that keyhole-level
    gains carry over to full navigation.
\end{myitem}

\section{Related Work}
\label{sec:related}

\noindent\textbf{Planning among Movable Obstacles.}
NAMO planners reason about obstacle relocations to enable
robot navigation~\cite{stilman2005namo,stilman2008artificial,vandenberg2009path,wu2010namo}.
Storage-zone planning assigns moved objects to storage
regions~\cite{ellis2023namo_storagezones}, while ReloPush
plans stable-pushing transfers to specified object goal
poses~\cite{ahn2025relopush}. Keyhole problems specify
robot reachability without assigning object destinations.
Interactive-FAR also connects local free-space regions
through stable-pushing motions~\cite{he2024interactivefar}.
ReloPush-BOSS optimizes intermediate poses of transferred
objects to enable paths to their goal
poses~\cite{ahn2026relopushboss}.
Haustein et al.~\cite{haustein2015kinodynamic} use a physics
model for nonprehensile transitions between resting
configurations. King et al.~\cite{king2016rearrangement}
combine object-centric actions with robot-centric motions
that permit simultaneous contact with multiple objects.
\name{} uses physics
simulation with learned priorities over push primitives
for keyhole opening.

\noindent\textbf{Learning to Guide Search.}
Learning can reduce planning computation through
abstraction selection, sampling guidance, or search
priorities. CAMPs~\cite{chitnis2021camps} learns
task-dependent constraints that permit removing irrelevant
state variables. NAMO-LLM~\cite{zhang2025namollm} uses a
pretrained language model to bias tree-node and obstacle
selection, while object target configurations remain
uniformly sampled. For greedy best-first search,
Garrett et al.~\cite{garrett2016ranking} formulate
heuristic learning as ranking states, aligning supervision
with the ordering used by search. \name{} applies this
ranking perspective to candidate push actions scored
from their parent states.

For pushing, MORE~\cite{huang2022interleaving} regresses
push-action values from Monte Carlo tree search (MCTS)
outcomes for object retrieval. Its predictions guide tree
expansion and rollouts, and its learned predictor is also
evaluated as a policy that selects pushes for execution.
\name{} likewise learns from search, but supplements value
regression with ranking supervision over push primitives
for keyhole opening.

\noindent\textbf{Learning Pushing Policies.}
Reinforcement learning has also been used to learn robot
velocity commands for keyhole
problems~\cite{yao2023local_pushable_rl} and pushing
commands with coordinated base-arm control and
obstacle-property estimation~\cite{yang2025namohrl}.
These methods train control policies from interaction
rewards. \name{} uses search outcomes to supervise rankings
over a supplied library of contact-and-duration pushes.
Whereas the evaluation in~\cite{yang2025namohrl}
treats object--wall collisions as failures, our setting
includes object--wall and object--object contacts that
affect push outcomes and the feasibility of subsequent
pushes.

\section{Problem Statement}
\label{sec:problem}

\begin{figure}[!tbp]
  \centering
  \includegraphics[width=\linewidth]{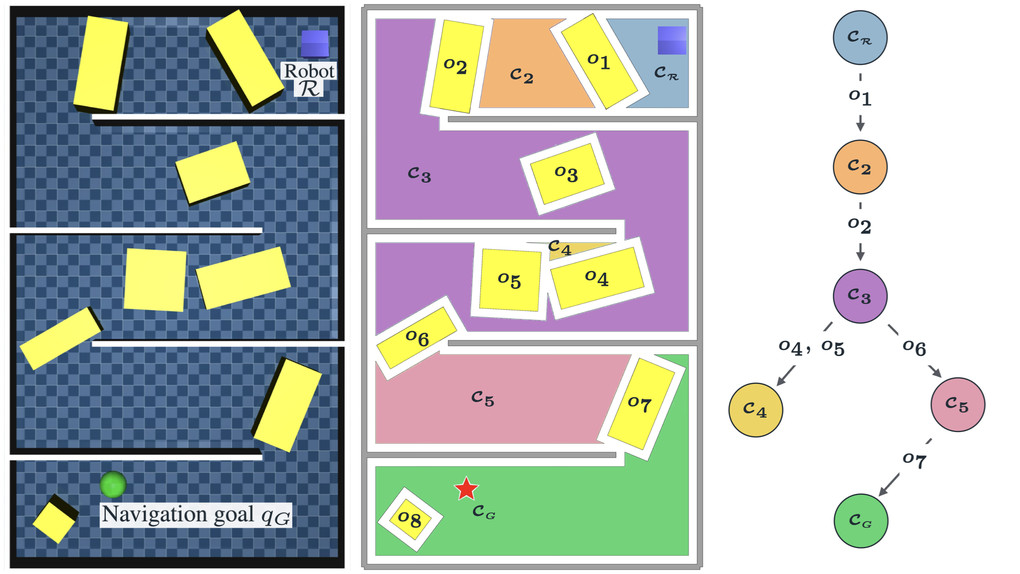}
    \caption{Illustrative simulated NAMO environment and its
    region graph. The robot starts at the top right, and the
    navigation goal lies at the bottom left. Graph vertices
    represent free-space regions, and edges identify movable
    objects separating neighboring regions.}
  \label{fig:problem_state}
\end{figure}

Consider a bounded planar workspace $\mathcal{W}$ containing
a mobile robot $\mathcal{R}$, $n$ box-shaped movable objects
$\mathcal{O}_m=\{o_1,\ldots,o_n\}$, and static obstacles
$\mathcal{O}_s$, such as walls (Fig.~\ref{fig:problem_state}).
The robot must navigate from its initial configuration
$q_{\mathcal{R}}$ to a goal $q_G$ without colliding with
static obstacles. When movable objects block all
collision-free paths, it must rearrange them through
non-prehensile pushing to reach the goal.

% {\bf NAMO Setup} 

We approximate the robot's footprint by a disk to determine
reachability in a planar configuration space
$\mathcal{C}\subset\mathbb{R}^2$. For the current object
arrangement, the collision-free configurations form
$\mathcal{C}_{free}\subseteq\mathcal{C}$, with connected
components $\mathcal{C}_{free}=\bigcup_{j=1}^{m}\mathcal{C}_j$.
Two components $\mathcal{C}_i$ and $\mathcal{C}_j$ are
\emph{neighboring} if a barrier of movable objects directly
separates them. A \emph{blocking set}
$\mathcal{B}(\mathcal{C}_i,\mathcal{C}_j)\subseteq\mathcal{O}_m$
contains one or more objects in that barrier whose
hypothetical removal connects the components.

Let $\mathcal{C}_{\mathcal{R}}$ be the \emph{source}
component containing the robot and $\mathcal{C}_i$ a selected
neighboring \emph{target} component. A keyhole
problem~\cite{stilman2005namo} asks the robot to rearrange
objects in $\mathcal{B}(\mathcal{C}_{\mathcal{R}},\mathcal{C}_i)$
so that its new reachable region $\mathcal{C}_{\mathcal{R}}'$
includes previously inaccessible configurations in the
initial target component. Displaced objects may occupy or
subdivide the original components, so
$\mathcal{C}_{\mathcal{R}}'$ need not contain either one
in its entirety. With $\mathcal{C}_G$ denoting the component
containing $q_G$, we solve successive keyhole problems
until the goal becomes reachable.

We consider keyhole problems in which the robot rearranges objects through non-prehensile pushing. Pushes may displace one object or multiple objects only through induced object--object contact.

\section{Proposed Method}
\label{sec:method}

% \name{} learns which pushes to evaluate first when solving
% a keyhole problem. We first define executable primitives
% and the search they guide, then describe how search outcomes
% supervise the ranker and how its scores support planning
% and direct execution.

\subsection{Primitive-Based Search}
\label{sec:search}
\label{sec:primitives}

\noindent\textbf{Push Primitives.}
\begin{wrapfigure}{R}{0.35\columnwidth}
  \centering
  \includegraphics[width=\linewidth]{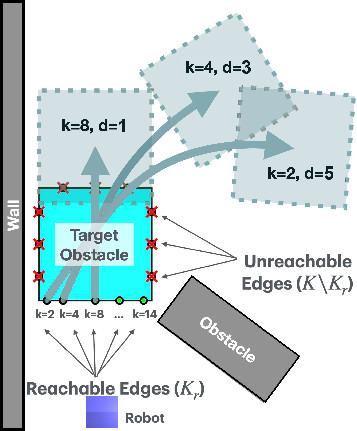}
  \caption{Push primitives at different contact points.}
  \label{fig:push_primitives}
\end{wrapfigure}
The action \texttt{PUSH}$(k,d)$ selects a contact point $k$
on an object face and a duration index $d$. The library
contains $K$ equally spaced contacts on each of the four
faces and $D$ duration choices. With time unit $t$, index
$d\in\{1,\ldots,D\}$ specifies a commanded duration of
$dt$, giving $P=4KD$ primitives per object.
The pushing direction is the inward normal of the contact
face at the start of the push and remains fixed in the
world frame. Face-center pushes translate the object,
while off-center pushes can also induce rotation
(Fig.~\ref{fig:push_primitives}). Contacts with surrounding
objects can alter these motions. To execute a primitive, the robot navigates to an approach
configuration offset outward from the contact point and
aligns its heading with the pushing direction. A
pure-pursuit controller tracks a straight line in this
direction for up to $dt$. Execution stops early if the
robot collides with a static obstacle or another movable
object, or if the pushed object stalls.
For reachability, obstacles are inflated by half the
diagonal of the robot's footprint. A contact is reachable
if its approach configuration is collision-free and
reachable from the robot's current configuration.
Primitives at such contacts are called reachable primitives.

% I rewrote above

\noindent\textbf{Search Procedure.}
\label{sec:search_procedure} For a source component $\mathcal{C}_{\mathcal{R}}$ and
target component $\mathcal{C}_i$, search considers sequences
of primitives applied to objects in
$\mathcal{B}(\mathcal{C}_{\mathcal{R}},\mathcal{C}_i)$.
A state $s$ specifies the poses of the robot and all movable
objects. Each call to the physics simulator applies a reachable primitive
to a selected object from $s$, with the bodies initially at
rest, and produces a successor $s'$. Trials causing robot
collisions with static obstacles or non-selected movable
objects are rejected.
A push that stalls after displacing the selected object can still produce a successor state. Reference configurations are uniformly sampled from the initial target component $\mathcal{C}_i$ and held fixed throughout
search. A state satisfies the keyhole success criterion
when the robot can reach at least a prescribed fraction of these
configurations. This tests access to the original target
region as object motion changes the free-space partition.

Greedy best-first search maintains a single priority queue
of candidate pushes across states. Each candidate specifies
its parent state, blocking object, and primitive, and a
supplied priority function orders the queue. At each
iteration, the highest-priority candidate is simulated from its parent
state. If the successor satisfies the success criterion,
search returns the sequence leading to it. Otherwise,
if the robot or pushed object has moved, reachable primitives
from the successor enter the queue, up to a maximum
sequence length of $L$ pushes. Search returns failure
when the queue is empty or its simulator-call budget is
exhausted.

\noindent\textbf{Computational Challenge.}
\label{sec:challenge}
With $b$ blocking objects and $P$ primitives per object,
each state admits at most $bP$ candidate pushes. Exhausting
the search tree through depth $L$ requires at most
$\sum_{\ell=1}^{L}(bP)^\ell = O((bP)^L)$ simulator calls.
For one supplied blocking object, the bound reduces to
$O(P^L)$. Moving one blocker can also change the reachability
and feasible motions of others.

A push may enable a later opening without itself satisfying
the keyhole success criterion. Identifying such pushes can
require evaluating their continuations. The objective of
learning is to rank candidate pushes from search outcomes
to reduce the simulator calls needed to find a keyhole
solution.

\subsection{Learning from Search}
\label{sec:learning}

% \name{} learns the priority function for the search in
% Sec.~\ref{sec:search_procedure}.
For a given keyhole problem,
let $a=(o,k,d)$ denote a push on object $o$ with contact $k$
and duration index $d$. The model, with parameters $\theta$,
assigns a score $h_\theta(s,a)$ to applying $a$ from state
$s$, with higher scores receiving earlier evaluation.

\noindent\textbf{Data Collection and Supervision.}
\label{sec:data}
Training data is collected through exhaustive
breadth-first search over sequences of up to two pushes
on a given blocking object, with each push evaluated
in simulation. For each initial state and blocking object,
a first push receives an immediate-success label if its
successor satisfies the keyhole criterion. Otherwise,
exhaustive search over second pushes determines whether
the first push is a \emph{setup push} that enables an opening.
Successor states are used only for lookahead. For training
observation $i$ with initial state $s_i$, each evaluated
first action $a$ receives target $y_{ia}=1$ for immediate
success, $y_{ia}=\gamma$ for a setup push, and $y_{ia}=0$ otherwise,
where $0<\gamma<1$ places immediate successes above setups.
Unreachable actions also receive zero targets.

\noindent\textbf{Model Architecture.}
\label{sec:model}
\begin{figure}[t]
\centering
\includegraphics[width=\columnwidth]{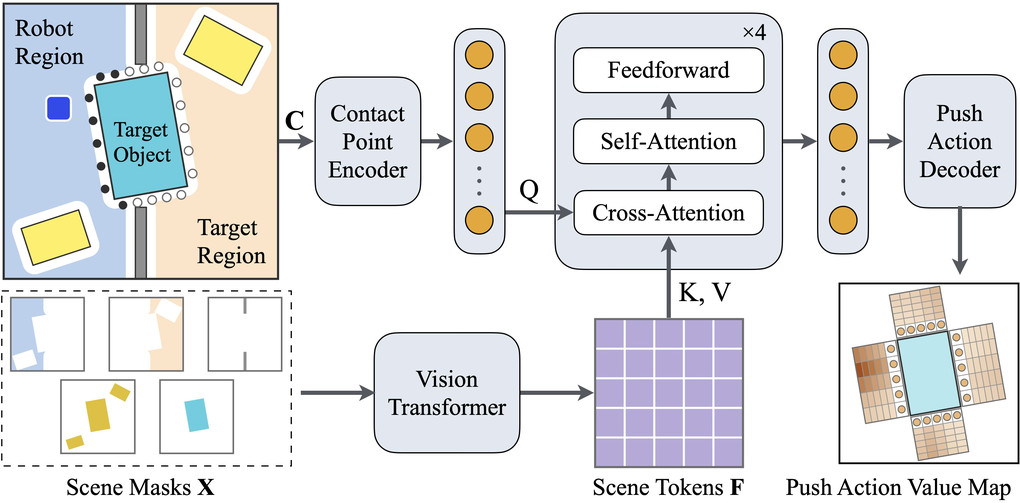}
\caption{\name{} architecture. The scene crop and contact
coordinates produce scores for contact-point and duration
choices. Blue and orange indicate the robot and target
regions, and filled markers indicate reachable contacts.
Only five contacts per face are shown, and scores are schematic.}
\label{fig:architecture}
\end{figure}
The model in Fig.~\ref{fig:architecture} associates
predictions with contact locations, following Spatial
Action Maps and HACMan~\cite{wu2020spatialaction,zhou2023hacman}.
For state $s$ and object $o$, it receives a crop
$\mathbf{X}\in[0,1]^{5\times H_{\mathrm{img}}\times W_{\mathrm{img}}}$
centered on $o$ and aligned with the world axes. The five
channels encode static obstacles, all movable objects,
the selected object, and the robot and target regions.
The region masks are recomputed from each state's geometry,
with the target mask using the same reference configurations
held fixed for the keyhole success test
(Sec.~\ref{sec:search_procedure}). The input also
includes the contact coordinates
$\mathbf{C}=(\mathbf{c}_1,\ldots,\mathbf{c}_{4K})$ in the crop.

% PROVENANCE (2026-09-14): The new NAMO archive explicitly identifies
% the reported runs as after the goal-picture fix. The timed frozen-600
% comparison retains its archived runs; do not relabel those as reruns.
% See audit/2026-09-14/REPORT.md for the separate evaluation protocols.

A vision transformer maps $\mathbf{X}$ to scene tokens
$\mathbf{F}$ using convolutional patch embeddings, learned
positional embeddings, and blocks of self-attention and
feedforward layers. The contact encoder forms each token
from three terms. Contact coordinates are normalized to
$[-1,1]^2$, Fourier encoded, and passed through an MLP.
A learned embedding identifies the object face and sample
index. A feature is sampled from $\mathbf{F}$ at the contact
by bilinear interpolation and passed through a linear layer.
The three terms are summed to initialize the token. Each contact block applies cross-attention with contact
tokens as queries and scene tokens as keys and values,
followed by self-attention over the contact tokens and
a feedforward layer. The push action decoder, shared
across contact tokens, estimates a score for each duration.
The outputs form a push action value map
over all $P$ primitives for the selected object in one
forward pass.

\noindent\textbf{Training Objective.}
\label{sec:training}
Greedy best-first search uses predicted scores to order
candidate pushes, motivating a ranking
objective~\cite{garrett2016ranking}.
The training objective combines a regression loss
$\mathcal{L}_{\mathrm{reg}}$ for targets $y_{ia}$ with ranking
losses that supervise comparisons within states.
% HL-Gauss description temporarily omitted at the author's request.
% Restore the following sentence and replace L_reg with L_cat throughout:
% The regression term $\mathcal{L}_{\mathrm{cat}}$ uses
% HL-Gauss~\cite{imani2018improving}.
Unreachable actions
receive reduced regression weight and remain zero-target
alternatives in the ranking terms. Within-state ranking uses two listwise terms: immediate
successes are ranked above setups and zero-target actions,
and setups above zero-target actions. For observation $i$
and level $\ell\in\{\gamma,1\}$, let
$\mathcal{P}_{i\ell}$ and $\mathcal{N}_{i\ell}$ contain
supervised actions with targets equal to $\ell$ and below
$\ell$, respectively. With $h_{ia}=h_\theta(s_i,a)$, the
loss for nonempty comparison sets is
\begin{equation}
\mathcal{L}_{i\ell}^{\mathrm{list}} =
-\frac{1}{|\mathcal{P}_{i\ell}|}
\sum_{a\in\mathcal{P}_{i\ell}}
\log
\frac{\exp(h_{ia}/\tau)}
{\sum_{b\in\mathcal{P}_{i\ell}\cup\mathcal{N}_{i\ell}}
 \exp(h_{ib}/\tau)},
\label{eq:within_state_loss}
\end{equation}
where $\tau>0$ is a temperature.
$\mathcal{L}_{\ell}^{\mathrm{list}}$ averages over observations
with both comparison sets nonempty and is zero if none exist.
The combined objective is
$\mathcal{L}=\mathcal{L}_{\mathrm{reg}}
+\lambda_1\mathcal{L}_{1}^{\mathrm{list}}
+\lambda_\gamma\mathcal{L}_{\gamma}^{\mathrm{list}}$,
where $\lambda_1$ and $\lambda_\gamma$ weight the two
applications of the within-state loss.
% Additional network and optimization settings remain in the
% commented experimental-setup block.

\subsection{Planning and Execution}
\label{sec:planning_execution}

\noindent\textbf{Search Guidance and Policy Execution.}
\label{sec:deployment}
Both deployment modes use the same trained model.
In \emph{\name{} Search}, $h_\theta(s,a)$ supplies the
priorities for the search in Sec.~\ref{sec:search_procedure}.
In \emph{\name{} Policy}, the scores select one push at
each decision, committing to its outcome without retaining
alternative branches.

On hardware, \name{} Policy executes the highest-scoring available
reachable push without simulating candidates. In simulation,
it evaluates candidates in score order and rejects pushes
that leave both the robot and pushed object poses unchanged.
If a trial reports a push failure, equal or longer durations
at the same contact on that object are excluded from the
same parent state. \name{} Policy commits the first accepted
successor. After each execution, the scene observation and
candidate set are updated before the next decision.

\noindent\textbf{NAMO Planning.}
\label{sec:namo_planning}
The NAMO planner constructs a region graph whose
vertices represent the free-space components defined
in Sec.~\ref{sec:problem}. Edges represent neighboring
component pairs and are labeled with their blocking
sets. Breadth-first search computes a shortest
route to $\mathcal{C}_G$ for each possible first region
transition. Each route receives one priority value per object blocking its first transition, equal to the number of region transitions plus that object’s prior attempts. The route with the smallest value is selected, after which the model compares reachable pushes across all eligible objects blocking its first transition.

For a keyhole with multiple blocking objects, the model
scores each eligible object separately, and the reachable
pushes across these objects are compared for search or
direct action selection.
\name{} Search composes local keyhole solutions in
simulation, updating the graph from each solution's
terminal state until $q_G$ becomes reachable or search
terminates. If a keyhole search exhausts its candidates
without a solution, the planner excludes the attempted
blocking objects for that region transition in the current
scene and considers another route. Search terminates when
no eligible route remains or an overall computation limit
is reached. \name{} Policy executes one push and selects
the next keyhole from the updated scene observation.

\FloatBarrier
\section{Experiments}
% Experiments evaluate whether learned push priorities reduce simulator calls
% and planning time in keyhole search, and whether retaining search alternatives
% changes success relative to policy execution. \namo{} experiments assess goal
% reaching through successive keyhole problems in simulation and on hardware.
\label{sec:experiments}

\subsection{Experimental Setup}
\label{sec:sim_setup}

\noindent\textbf{Environments.}
Simulation experiments use MuJoCo~\cite{todorov2012mujoco}
with a differential-drive robot and box-shaped movable objects.
Test scenes reuse the training wall templates with unseen
configurations of object poses and dimensions.

\noindent\textbf{Training.}
Approximately 1.3 million observations are split 90/10
for training and validation by source scene.
The objective uses $\gamma=0.5$, $\tau=0.15$,
$(\lambda_1,\lambda_\gamma)=(0.1,0.05)$, and
unreachable-action regression weight $0.1$.
Network architecture and optimization settings are
provided on the project page.

\noindent\textbf{Comparison methods.}
\name{} Search, Random Search, and Geometric Search use
the same primitive library and search procedure
(Sec.~\ref{sec:search}), differing only in their priorities.
\name{} Search uses learned scores. Random Search assigns
independent random priorities to reachable candidates at
each generated state. Geometric Search uses free-space pose
transformations for each primitive on three reference boxes:
wide, tall, and square. It selects the reference box with
the closest aspect ratio, applies its primitive transformation
to the selected object's current pose, and scores target-region
reachability after virtually placing the object at that pose.

% For the main comparisons, all search methods consider sequences
% of at most two pushes per keyhole problem, matching the training
% horizon. This choice reflects an inherent efficiency trade-off
% between exploring alternative plans and extending a single
% trajectory. With $b$ candidates per state, depth-$d$ search can
% retain $O(b^d)$ frontier entries in the worst case, and exhaustive
% exploration incurs $O(b^d \log(b^d))$ queue-management time with
% a binary heap, in addition to simulator evaluations.
% In contrast, a greedy $d$-push rollout considers $O(bd)$
% candidates while retaining only the current candidate pool.
% We therefore compare bounded-depth search with longer
% \name{} Policy rollouts, contrasting the exploration of
% alternative short plans with the continuation of a single
% greedy trajectory while keeping the learned model fixed.
% The ablations in Sec.~\ref{sec:policy_search} empirically
% examine this trade-off between search depth and computational
% cost.

\noindent\textbf{Planning parameters.}
The library uses $K=15$ contacts per face and $D=5$
durations with unit $t=1.0\,\mathrm{s}$, giving
$P=4KD=300$ primitives per object.
The search depth limit $L$ specifies the maximum
number of pushes in a candidate plan.

\noindent\textbf{Metrics.}
Keyhole success requires at least $20\%$ of up to 100
reference configurations fixed in the initial target
component to become reachable.
A simulator call evaluates one push. Planning time is
measured as wall-clock search time on the same machine
for all methods. SR@$B$ is the percentage of test problems
solved within $B$ simulator calls.

%  \textbf{Learned Model and Training.}
% Simulated searches on keyhole problems provide approximately
% 1.3 million observations and their associated action outcomes
% (Sec.~\ref{sec:data}). The dataset uses a 90/10
% training--validation split by source scene.
%
% The five-channel input uses an object-centered
% $0.5\,\mathrm{m}\times0.5\,\mathrm{m}$ crop at
% $64\times64$ resolution. Image patches are $4\times4$.
% The scene and contact stacks each contain four blocks
% with token dimension 192, six attention heads, and
% feedforward width 768. Contact coordinates use eight
% Fourier frequencies per coordinate. Each contact token
% predicts $J=51$ score bins per duration.
% Renderer/evaluation provenance is separated by study in audit/2026-09-14/REPORT.md.
%
% Models are trained for 12 epochs using
% AdamW with base learning rate $3\times10^{-4}$, linear
% warmup followed by cosine decay, weight decay $0.01$,
% and batches of up to 256 observations grouped
% by source scene and blocking object. Checkpoint selection minimizes the
% validation categorical loss. The objective in
% Sec.~\ref{sec:training} uses $\gamma=0.5$,
% $\sigma=0.75/J$, $\eta=0.1$, $\tau=0.15$,
% and $(\lambda_1,\lambda_\gamma)=(0.1,0.05)$.

\subsection{Keyhole Search in Simulation}
\label{sec:ro_results_sim}
% Selected single-column layout: each distribution above its table.
% Distribution labels are abbreviated; tables and caption retain method context.
\begin{figure}[!t]
    \centering
    \begingroup
    \footnotesize
    \setlength{\tabcolsep}{1.5pt}
    \renewcommand{\arraystretch}{1.05}

    \makebox[\linewidth][l]{%
        \textbf{(a) Simulator calls}%
    }
    \par\vspace{3pt}

    \includegraphics[width=\linewidth]{%
        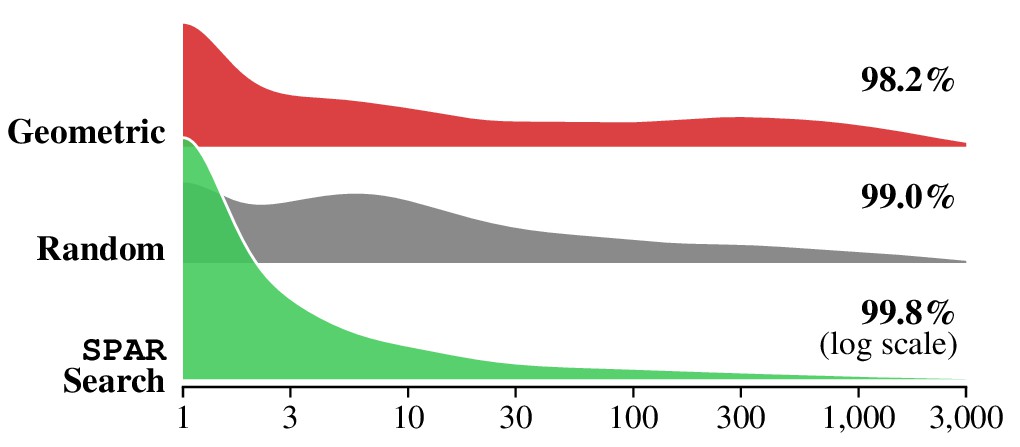}
    \par\vspace{4pt}
    \begin{tabular*}{\linewidth}{@{\extracolsep{\fill}}l*{8}{c}@{}}
    \toprule
    & \multicolumn{4}{c}{One Push}
    & \multicolumn{4}{c}{Two Pushes} \\
    \cmidrule(lr){2-5} \cmidrule(lr){6-9}
    Method
        & \textbf{Overall} & Easy & Medium & Hard
        & \textbf{Overall} & Easy & Medium & Hard \\
    \midrule
    \shortstack[l]{Geometric}
        & 2 & \textbf{1} & 3 & 6.5
        & 53 & 10 & 60 & 287 \\
    \shortstack[l]{Random}
        & 4 & \textbf{1} & 4 & 9
        & 35 & 9.5 & 37 & 305 \\
    \shortstack[l]{\name{} Search}
        & \textbf{1} & \textbf{1} & \textbf{1} & \textbf{2}
        & \textbf{5} & \textbf{3} & \textbf{5} & \textbf{16} \\
    \bottomrule
\end{tabular*}

    \par\vspace{8pt}

    \makebox[\linewidth][l]{%
        \textbf{(b) Planning time (s)}%
    }
    \par\vspace{3pt}

    \includegraphics[width=\linewidth]{%
        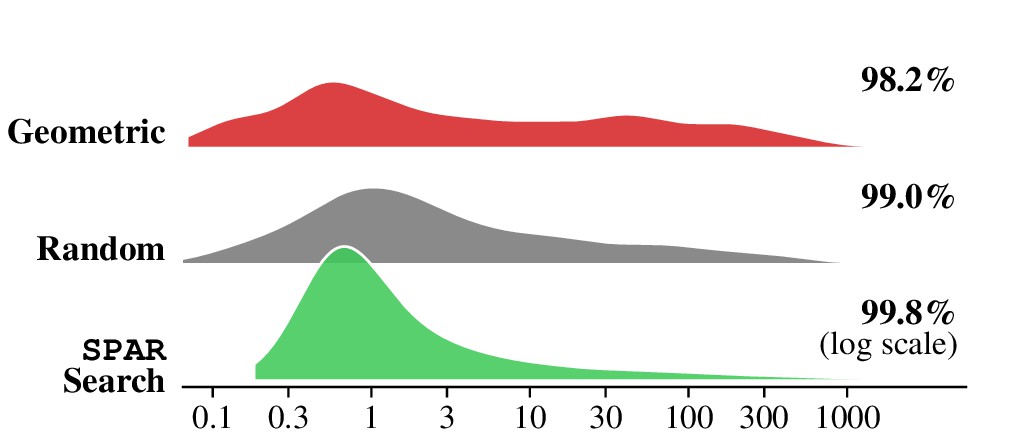}
    \par\vspace{4pt}
    
\begin{tabular*}{\linewidth}{@{\extracolsep{\fill}}l*{8}{c}@{}}
    \toprule
    & \multicolumn{4}{c}{One Push}
    & \multicolumn{4}{c}{Two Pushes} \\
    \cmidrule(lr){2-5} \cmidrule(lr){6-9}
    Method
        & \textbf{Overall} & Easy & Medium & Hard
        & \textbf{Overall} & Easy & Medium & Hard \\
    \midrule
    \shortstack[l]{Geometric}
        & \textbf{0.52} & \textbf{0.23} & \textbf{0.57} & 1.47
        & 11.88 & 1.98 & 12.60 & 63.02 \\
    \shortstack[l]{Random}
        & 0.74 & 0.31 & 0.85 & 1.68
        & 6.56 & 1.96 & 7.91 & 64.09 \\
    \shortstack[l]{\name{} Search}
        & 0.57 & 0.45 & 0.59 & \textbf{0.68}
        & \textbf{1.72} & \textbf{1.18} & \textbf{1.70} & \textbf{5.10} \\
    \bottomrule
\end{tabular*}

    \par
    \endgroup

    \input{figures/keyhole/caption}
\end{figure}

The benchmark contains 300 problems requiring one push
and 300 requiring two, with 100 per difficulty tier in
each group. Each has a fixed set of one or more blocking
objects. Searches use $L=2$ and a budget of 3,000 simulator
calls per problem. Difficulty is the expected position of the first opening
push (one-push problems) or setup push (two-push problems)
in a uniformly random ordering of candidates.
A lower expected position means fewer trials, on average,
to find an opening or setup push. Such a push is easier to
find when more reachable pushes can lead to a solution,
or when there are fewer reachable pushes to choose from.
Easy, medium, and hard correspond to expected positions
of at most $3$, above $3$ and at most $15$, and above $15$,
respectively. For two-push problems, this tests whether
search can find setup pushes when they are scarce.
Success rates average over all attempts, while cost medians
pool successful trials from five Random Search repetitions,
three \name{} training seeds, and one Geometric Search run
per problem. Reported speedups are ratios of these
success-conditioned medians.

\noindent\textbf{Learning Speeds Up Success.}
Despite scoring overhead, \name{} Search achieves
$1.3\times$ and $3.8\times$ speedups in overall median
wall-clock time relative to Random Search on one-push
and two-push problems, respectively
(Fig.~\ref{fig:region_opening_results}).
On hard two-push problems, this advantage reaches
$12.6\times$, although Random Search remains faster
on easy one-push problems.
Geometric Search is a strong baseline on one-push
problems, achieving the lowest overall median
wall-clock time.
Even in this setting, however, \name{} Search achieves
a $2.2\times$ speedup on the hard subset.
Its advantage over Geometric Search becomes more
pronounced on two-push problems, with a $6.9\times$
speedup in overall median wall-clock time.

\noindent\textbf{Multi-Push Problems Demand Foresight.}
Two-push search admits up to $300^2$ candidate sequences
for a single blocking object. This motivates guidance that
identifies setup pushes by the openings they enable next,
even when they do not open the target region themselves.
Geometric Search offers little guidance for choosing such
pushes because it scores only immediate target-region
reachability at the primitive's nominal endpoint.
Random Search explores candidate sequences without guidance.
On these problems, \name{} Search reduces the overall median
simulator cost by $7.0\times$ relative to Random Search and
$10.6\times$ relative to Geometric Search
(Fig.~\ref{fig:region_opening_results}).

\noindent\textbf{Difficulty Tiers Track Search Cost.}
Across easy, medium, and hard tiers, Random Search's median
simulator cost increases from $1$ to $4$ to $9$ calls on
one-push problems and from $9.5$ to $37$ to $305$ on two-push
problems (Fig.~\ref{fig:region_opening_results}).
These trends support the benchmark stratification: the tiers
reflect differences in the cost of finding solutions, even
though the two-push metric excludes continuation search.

\subsection{Generalization Beyond Two Pushes}
\label{sec:longer_keyholes}

\begin{figure}[!tbp]
    \centering
    \includegraphics[width=\columnwidth]{%
        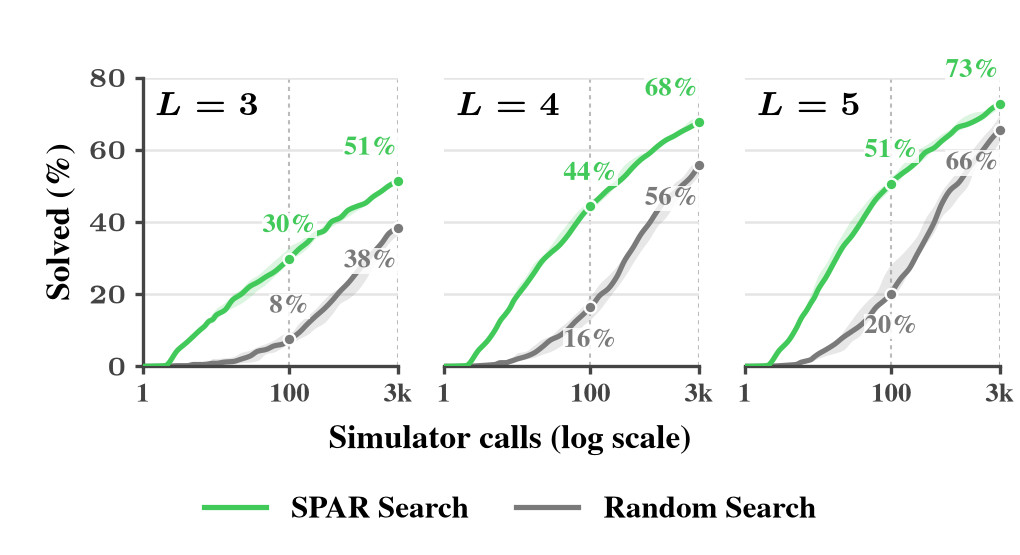}
    \caption{Keyhole success versus simulator calls at depth
    limits $L\in\{3,4,5\}$, measured over all 600 problems.
    Lines show median success across three \name{} training
    seeds or five Random Search evaluation seeds. Shading
    spans the minimum to maximum.}
    \label{fig:longer_keyholes}
\end{figure}

This experiment evaluates whether priorities learned from
searches of at most two pushes on a given blocking object
guide search at greater depths. The benchmark contains
600 keyhole problems, each verified by exhaustive search
to have no solution within two pushes under the primitive
library. $69$ of them are unknown to have a solution under our primitive library. \name{} Search and Random Search
are compared on the same problems at depth limits $L\in\{3,4,5\}$,
with a budget of 3,000 simulator calls per problem.
The model is used without retraining. \name{} Search solves more problems than Random Search
at matched depths and simulator-call budgets
(Fig.~\ref{fig:longer_keyholes}). At $L=5$, it solves
$50.7\%$ within 100 calls, compared with $20.2\%$ for Random
Search. Increasing $L$ from three to five raises success
at 3,000 calls from $51.3\%$ to $72.8\%$ for \name{} Search
and from $38.3\%$ to $65.5\%$ for Random Search.
Increasing the depth limit admits longer plans, while
the gains at matched depths show that priorities learned
from two-push outcomes continue to guide search well beyond
the training horizon.

\subsection{\namo\ in Simulation}
\label{sec:namo_results_sim}
The benchmark contains 300 \namo{} environments, with
100 per difficulty tier. Each initial shortest route
contains two region transitions. Keyholes may be
independent, or object motions at one may affect the next.
\name{} Search and Random Search use the \namo~planner in
Sec.~\ref{sec:namo_planning}, with $L=2$ per keyhole and
a budget of 9,000 simulator calls per \namo{} problem. Our evaluation focuses on how efficiently the methods find solutions in environments known to be solvable by our primitive library. The benchmark comprised environments that are solved by at least one of five Random Search runs under this budget.
Their median call count, with failures assigned the budget,
defines easy ($<30$), medium ($[30,300)$), and hard
($\geq300$) tiers. Evaluation uses five Random Search seeds
independent of those that defined the difficulty tiers.

\begin{figure}[!tbp]
    \centering
    \includegraphics[width=\columnwidth]{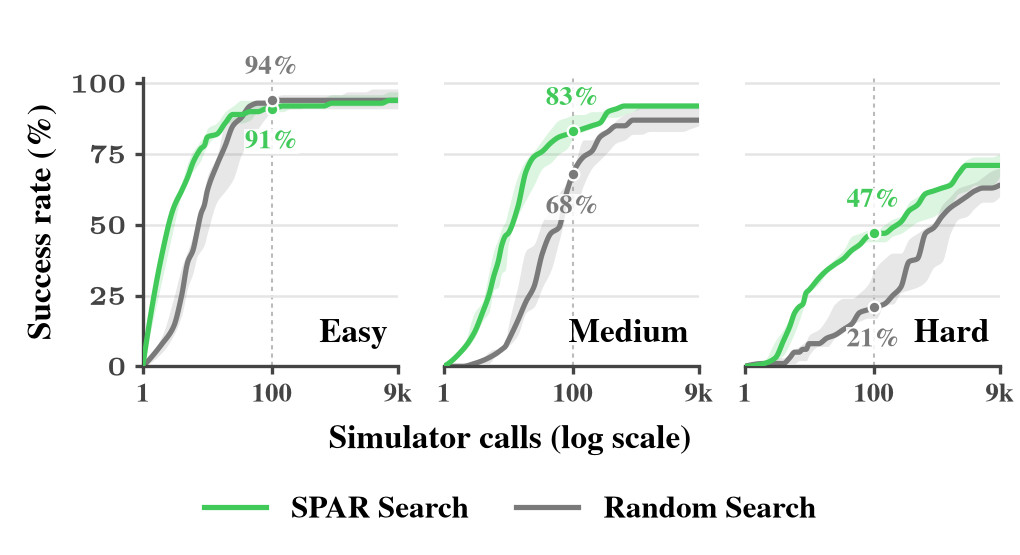}
    \caption{\namo{} success rate versus simulator-call budget.
    Lines show median success across three \name{} training
    seeds or five Random Search evaluation seeds, with shading
    spanning the minimum to maximum. Curves and bands are
    smoothed for display after aggregation.}
    \label{fig:full_namo_budget}
\end{figure}

\noindent\textbf{Keyhole Efficiency Transfers to \namo.}
Learned keyhole priorities improve \namo{} success
at limited simulator-call budgets
(Fig.~\ref{fig:full_namo_budget}).
At $100$ calls, all three \name{} Search seeds achieve
higher success rates than every Random Search seed on
medium and hard scenes. These differences narrow by the
$9{,}000$-call limit, but median success on hard scenes
remains $71\%$ for \name{} Search versus $64\%$ for
Random Search.

% In 18 of the 300 environments, \name{} Search reaches
% the goal after exhausting at least one local keyhole
% search.

\subsection{Ablation Studies}
\label{sec:ablations}

Table~\ref{tab:ablations} compares the full \name{} Search
model with four training and architecture variants.
All ablations use the 600-problem keyhole benchmark
with depth limit $L=2$ and a budget of $B=3{,}000$
simulator calls per problem.
% The ablations evaluate how the training objective and contact
% representation affect search effort. Each variant is evaluated on
% the same 1,328 one-push and 992 two-push keyhole problems, each
% with one blocking object. Search uses a horizon of $L=2$ and a
% budget of 900 simulator evaluations. This population
% differs from the 1,310 one-push and 973 two-push problems in
% Table~\ref{tab:region_opening_costs}, which uses a 4,000-evaluation
% budget. Each learned variant uses three training seeds. Success
% at a budget is the fraction of problems solved within that number
% of simulator evaluations, including unsolved problems in the
% denominator.

\begin{table}[!tbp]
    \centering
    \caption{Keyhole success (SR@$B$, \%). Entries report
    mean $\pm$ sample standard deviation
    over three training seeds.}
    \label{tab:ablations}
    \footnotesize
    \setlength{\tabcolsep}{2pt}
    \begin{tabular}{@{}lccc@{}}
        \toprule
        & One-push & \multicolumn{2}{c}{Two-push} \\
        \cmidrule(lr){2-2}\cmidrule(lr){3-4}
        Variant & SR@1 & SR@5 & SR@30 \\
        \midrule
        Full model (\name{} Search) & $74.4 \pm 2.4$ & $51.7 \pm 1.5$ & $78.9 \pm 0.8$ \\
        \midrule
        \multicolumn{4}{@{}l}{\textit{(a) Loss function}} \\
        Regression only & $72.6 \pm 1.6$ & $43.9 \pm 0.4$ & $74.4 \pm 1.3$ \\
        No unreachable supervision & $73.6 \pm 1.0$ & $43.9 \pm 2.2$ & $73.9 \pm 1.7$ \\
        \midrule
        \multicolumn{4}{@{}l}{\textit{(b) Architecture}} \\
        Global readout & $59.2 \pm 1.0$ & $43.8 \pm 1.8$ & $73.9 \pm 2.0$ \\
        No contact self-attention & $71.9 \pm 0.4$ & $49.8 \pm 1.0$ & $79.0 \pm 0.3$ \\
        \bottomrule
    \end{tabular}
\end{table}

\noindent\textbf{Training Objective.}
\emph{Regression only} removes the ranking losses. \emph{No unreachable
supervision} removes the zero-target supervision assigned
to unreachable actions. Both variants reduce success
more on two-push problems than on one-push problems at
the reported budgets.
% The September 7 implementation audit and September 15 manuscript
% revision differ on unreachable-action ranking membership. The definition
% above states the shared intervention without attributing its effect to
% regression alone; verify the evaluated training configuration separately.

\noindent\textbf{Contact Representation.}
\emph{Global readout} replaces the contact-token pathway
with a scene-level head that predicts all primitive scores.
\emph{No contact self-attention} removes attention among
contact tokens while retaining cross-attention to scene
tokens. The global readout reduces success at every
reported budget, supporting the contact-token representation.
Removing self-attention produces smaller reductions
at SR@1 and SR@5, with essentially unchanged SR@30.

% \textbf{Contact representation.}
% The architecture ablations modify Sec.~\ref{sec:model}.
% \emph{Global readout} replaces the contact-token pathway with
% a scene-level readout that predicts scores for all primitives.
% This reduces one-push success within one evaluation from $82.5\%$
% to $58.2\%$ and two-push success within five evaluations from
% $64.9\%$ to $42.5\%$. Three further variants separately remove
% self-attention among contact tokens, the contact-index embedding,
% or the bilinearly sampled scene feature.
% Their two-push success rates within five evaluations are
% $59.4\%$, $63.0\%$, and $63.6\%$, respectively. Removing local
% feature sampling leaves mean one-push success within one
% evaluation at $82.5\%$.

% At 900 simulator evaluations, mean two-push
% success ranges from $92.4\%$ to $93.5\%$ across the learned
% variants, compared with $93.1\%$ for the full model and $88.4\%$
% for Random. Thus, the differences at five evaluations primarily
% reflect how quickly search finds a solution within this protocol.

\noindent\textbf{Search versus Policy.}
\label{sec:policy_search} The two deployment modes in
Sec.~\ref{sec:deployment} use the same trained model:
\name{} Search explores alternative sequences with $L=2$,
while \name{} Policy can extend a single trajectory beyond
two pushes.
For \name{} Policy, simulator calls execute
attempted pushes in the simulation environment.

\begin{figure}[!tbp]
    \centering
    \includegraphics[width=\columnwidth]{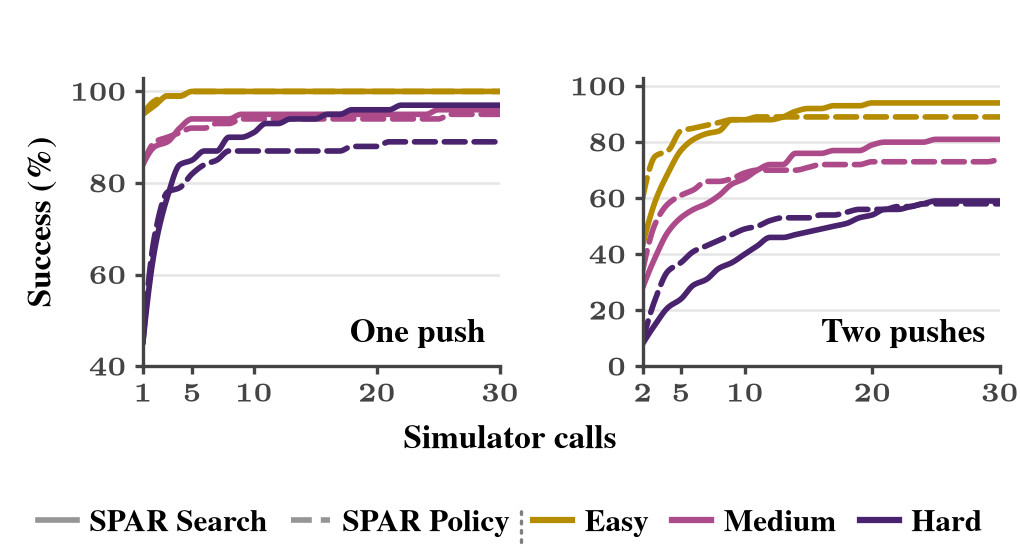}
    \caption{Keyhole success rate versus simulator-call budget.
      Solid lines show \name{} Search and dashed lines show
      \name{} Policy; colors denote difficulty. Each curve shows
      the median success rate across three training
      seeds. Curves are smoothed for display after aggregation.}
    \label{fig:policy_search}
\end{figure}

\name{} Policy initially solves more two-push problems in every
difficulty group (Fig.~\ref{fig:policy_search}). On easy
two-push problems, median success at two calls is $60\%$ for \name{} Policy
and $43\%$ for Search. Committing to a successor lets \name{} Policy
evaluate its continuation on the next call, whereas Search
may evaluate candidates from other states.

Across the three training seeds, increasing \name{} Policy's
budget from $30$ to $3{,}000$ simulator calls raises the median
number of solved problems from $505/600$ to $526/600$, a gain
of only $21$ successes. At $3{,}000$ calls, the median for \name{} Search is $599/600$, using sequences of at most two pushes. On this benchmark, retaining alternative sequences therefore yields higher
success than extending a single policy trajectory.
Policy may reach states with no feasible continuation,
whereas Search can revisit earlier simulated states.

\FloatBarrier
\subsection{Real-Robot \namo\ Experiments}
\label{sec:namo_results_real}

% Declare the results float early so it can share the page with the setup.
\begin{figure}[!t]
    \centering

    \includegraphics[width=\columnwidth]{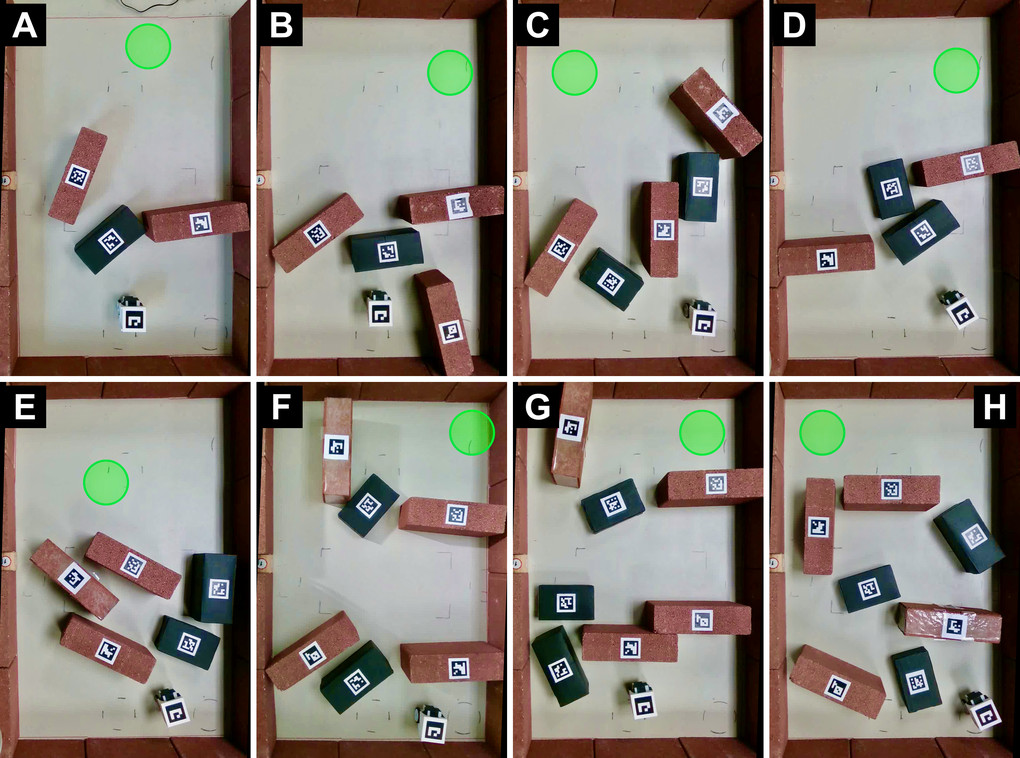}

    \vspace{0.5em}

    \footnotesize
    \setlength{\tabcolsep}{2pt}
    \ADLactivate
    \setlength{\dashlinedash}{0.4pt}
    \setlength{\dashlinegap}{1.5pt}
    % Center the divider between the right-aligned Search calls and Policy fractions.
    \settowidth{\dimen0}{Succ.}
    \settowidth{\dimen2}{5/5}
    \edef\hardwareDividerPadding{\the\dimexpr(\dimen0-\dimen2)/2\relax}
    % arydshln must also count booktabs' rules and padding in the divider height.
    \makeatletter
    \newcommand{\hardwareRuleSpace}[1]{\noalign{\adl@hline\adl@connect{#1}}}
    \makeatother
    % Split the horizontal rules at the Search/Policy gutter, preserving their padding.
    \newcommand{\hardwareSplitRule}[3]{%
        \noalign{\vskip\dimexpr#2-\aboverulesep\relax}%
        \cmidrule[#1](r{4pt}){1-7}\cmidrule[#1](l{2pt}){8-9}%
        \noalign{\vskip\dimexpr#3-\belowrulesep\relax}%
        \hardwareRuleSpace{\dimexpr#2+#1+#3\relax}%
    }
    \begin{tabular*}{\columnwidth}{@{\extracolsep{\fill}}l*{2}{crr}:@{\hspace{\hardwareDividerPadding}\extracolsep{2pt}}c@{\hspace{4pt}\extracolsep{\fill}}r@{}}
        \hardwareSplitRule{\heavyrulewidth}{\abovetopsep}{\belowrulesep}
        & \multicolumn{3}{c}{Random Search}
        & \multicolumn{3}{c:}{\name{} Search}
        & \multicolumn{2}{c}{\name{} Policy} \\
        \cmidrule(lr){2-4}\cmidrule(lr){5-7}\cmidrule(lr){8-9}
        \hardwareRuleSpace{\dimexpr\aboverulesep+\cmidrulewidth+\belowrulesep\relax}
        Scene & Succ. & \shortstack{Pushes} & \shortstack{Sim calls} & Succ. & \shortstack{Pushes} & \shortstack{Sim calls} & Succ. & \shortstack{Pushes} \\
        \hardwareSplitRule{\lightrulewidth}{\aboverulesep}{\belowrulesep}
        % Costs are medians over successful trials in real_exp_complete_envs.csv.
        % Scene mapping: B=shm_002_v9, C=shm_001_v2.
        A & 5/5 & 4 & 114 & 5/5 & 3 & 7 & 5/5 & 3 \\
        B & 5/5 & 1 & 9 & 5/5 & 1 & 3 & 5/5 & 1 \\
        C & 1/5 & 2 & 212 & 2/5 & 2.5 & 249 & 3/5 & 3 \\
        D & 1/5 & 7 & 791 & 5/5 & 4 & 115 & 5/5 & 4 \\
        E & 2/5 & 2 & 480.5 & 5/5 & 2 & 28 & 5/5 & 2 \\
        F & 5/5 & 4 & 38 & 5/5 & 3 & 9 & 5/5 & 5 \\
        G & 5/5 & 6 & 189 & 2/5 & 4 & 14 & 5/5 & 7 \\
        H & 3/5 & 6 & 34 & 4/5 & 5 & 48.5 & 4/5 & 6 \\
        \hardwareSplitRule{\lightrulewidth}{\aboverulesep}{\belowrulesep}
        Total & \multicolumn{3}{c}{27/40}
        & \multicolumn{3}{c:}{33/40}
        & \multicolumn{2}{c}{\textbf{37/40}} \\
        \hardwareSplitRule{\heavyrulewidth}{\aboverulesep}{\belowbottomsep}
    \end{tabular*}
    \caption{Real-robot NAMO environments and results.
Black objects are movable, red bricks form static obstacles
and walls, and green circles mark the navigation goal.
Success counts are out of five trials per method and scene.
Physical push attempts and simulator calls are reported as
medians over successful trials only.}
    \label{fig:real_robot_results}
\end{figure}

\noindent\textbf{Setup.}
Experiments compare \name{} Search, \name{} Policy, Random Search, and a navigation baseline on eight
environments, with five trials per method and environment.
The differential-drive robot~\cite{yu2017portable} has
the same footprint as its simulated model. ArUco tags
provide robot and object poses. These poses and measured
object dimensions are used to construct the MuJoCo scenes
for planning. Only the robot's speed is matched to
the hardware under duration commands. Contact
dynamics are not calibrated. We evaluate sim-to-real transfer of the learned priorities on tasks known to be physically solvable under our primitive library; scenes are retained if at least one Random Search screening run succeeds on hardware. These runs are separate from the reported trials. The navigation baseline ignores movable objects
when planning, then attempts to follow the route with
the objects present.

\noindent\textbf{Execution.}
\name{} Search and Random Search compute a plan in
simulation and execute its first push. After each attempt,
the scene is re-observed. If the scene has not changed, we discard the contact point from further execution. The remaining plan is reused
only if its next contact is reachable and re-simulation
confirms that the plan remains valid from the observed
state. Otherwise, planning restarts from the observed state,
excluding contacts that produced insufficient object
motion. Each fresh planning call uses depth limit $L=2$ per
keyhole. Trials are capped at 900 simulator calls.
\name{} Policy executes the highest-scoring reachable push
without simulating candidates, then selects the next action
from the observed state.
Once the goal becomes reachable, the robot navigates to it.
A trial ends after a total of 20 physical push attempts or
20 replanning attempts, or earlier if no executable subgoal remains.

% The author corrected the physical push limit to 20 attempts per trial
% on 2026-09-15, superseding the earlier no-cap statement. Unsuccessful
% physical attempts count toward this limit. The 900-call allowance
% remains per fresh planning call.
% The author confirmed on 2026-09-17 that the 20-replanning limit
% also applies across the whole trial, without resetting after a push.
% Saved scene D records count remaining-plan verification outside this
% allowance; its scope across the other trial configurations is unverified.

\noindent\textbf{Metrics.}
Task success requires the robot to reach the navigation goal
within a $5\,\mathrm{cm}$ position tolerance.
Simulator calls count push evaluations during search and
verification of a remaining plan. Physical pushes count attempts
executed on hardware, including those with insufficient object
displacement. Fig.~\ref{fig:real_robot_results} reports the results. Trial duration is
omitted because battery levels can limit the robot's velocity.

\noindent\textbf{Learning Improves Plan Discovery.}
\name{} Search finds an initial simulated plan in all $40$
trials, compared with $31/40$ for Random Search. The nine
remaining Random Search trials terminate before any physical
push. However, finding a simulated plan does not ensure task
completion: \name{} Search reaches the navigation goal in
$33/40$ trials.

\noindent\textbf{Learned Priorities Transfer Beyond Simulation.}
Trained entirely in simulation, \name{} Policy completes
$37/40$ hardware trials without hardware-specific contact
calibration, compared with $33/40$ for \name{} Search
and $27/40$ for Random Search. The navigation baseline
completes only $2/40$ trials, both in Scene D.
The Policy results show that priorities learned in simulation
support closed-loop \namo{} execution without online push simulation.
Search requires a simulated plan before execution, whereas
Policy selects pushes without that requirement.
Observed Search failures included simulated plans that did
not transfer to hardware, leaving the robot stuck, and failures
to find a plan within the search budget.
Scene G highlights the distinction between plan discovery
and physical success: \name{} Search finds an initial
plan in all five trials but completes two, while
\name{} Policy and Random Search each complete all five.
In Scene G, pushes landed differently from simulated predictions,
and Search failed to find continuations after physical pushes.
Among successful trials, \name{} Policy's median push count
is within one attempt of \name{} Search in six of the eight
scenes, and two to three attempts higher in the remaining two.

\FloatBarrier

\section{Limitations \& Future Work}
% \name{} Search and the \namo{} pipeline open several avenues for future exploration.
Our environments consist only of box-shaped objects with uniform mass and friction. Richer object geometries or contact properties may require more expensive search and adaptation to physical variability. The NAMO planner ranks routes by region-transition count and
previous attempts, without estimating the effort required to
open each keyhole. It commits to keyhole solutions without
backtracking over earlier openings, which may restrict access
to subsequent keyholes. Integrating learned keyhole search
with planners that account for relocation order and future
accessibility remains future work.

% Uncomment and replace for the camera-ready version, if applicable.
% \section*{Acknowledgment}
% Add acknowledgments and funding information here.

\bibliographystyle{IEEEtran}
\bibliography{references}

\end{document}